\documentclass[times,twocolumn,final,authoryear]{elsarticle}

\usepackage{prletters}
\usepackage{framed,multirow}
\usepackage[table,xcdraw]{xcolor}
\usepackage{booktabs}

\usepackage{amssymb}
\usepackage{latexsym}

\usepackage{url}
\usepackage{xcolor}
\definecolor{newcolor}{rgb}{.8,.349,.1}

\journal{Pattern Recognition Letters}

\begin{document}

\begin{frontmatter}

\title{Visual Question Answering: which investigated applications?}

\author[1]{Silvio \snm{Barra}\corref{cor1}} 
\cortext[cor1]{Corresponding author: 
  }
\ead{silvio.barra@unina.it}
\author[2]{Carmen  \snm{Bisogni}}
\author[3]{Maria  \snm{De Marsico}}
\author[4]{Stefano \snm{Ricciardi}}

\address[1]{University of Naples Federico II, Naples, Italy}
\address[2]{University of Salerno, Salerno, Italy}
\address[3]{Sapienza University of Rome, Rome, Italy}
\address[4]{University of Molise, Campobasso, Italy}

\received{1 May 2013}
\finalform{10 May 2013}
\accepted{13 May 2013}
\availableonline{15 May 2013}
\communicated{S. Sarkar}

\begin{abstract}
Visual Question Answering (VQA) is an extremely stimulating and challenging research area where Computer Vision (CV) and Natural Language Processig (NLP) have recently met. In image captioning and video summarization, the semantic information is completely contained in still images or video dynamics, and it has only to be mined and expressed in a human-consistent way. Differently from this, in VQA semantic information in the same media must be compared with the semantics implied by a question expressed in natural language, doubling the artificial intelligence-related effort. Some recent surveys about VQA approaches have focused on methods underlying either the image-related processing or the verbal-related one, or on the way to consistently fuse the conveyed information. Possible applications are only suggested, and, in fact, most cited works rely on general-purpose datasets that are used to assess the building blocks of a VQA system. This paper rather considers the proposals that focus on real-world applications, possibly using as benchmarks suitable data bound to the application domain. The paper also reports about some recent challenges in VQA research.
\end{abstract}


\end{frontmatter}


\section{Introduction}
\label{s:intro}

\begin{figure*} [ht!]
    \centering
    \includegraphics[width=0.8\textwidth]{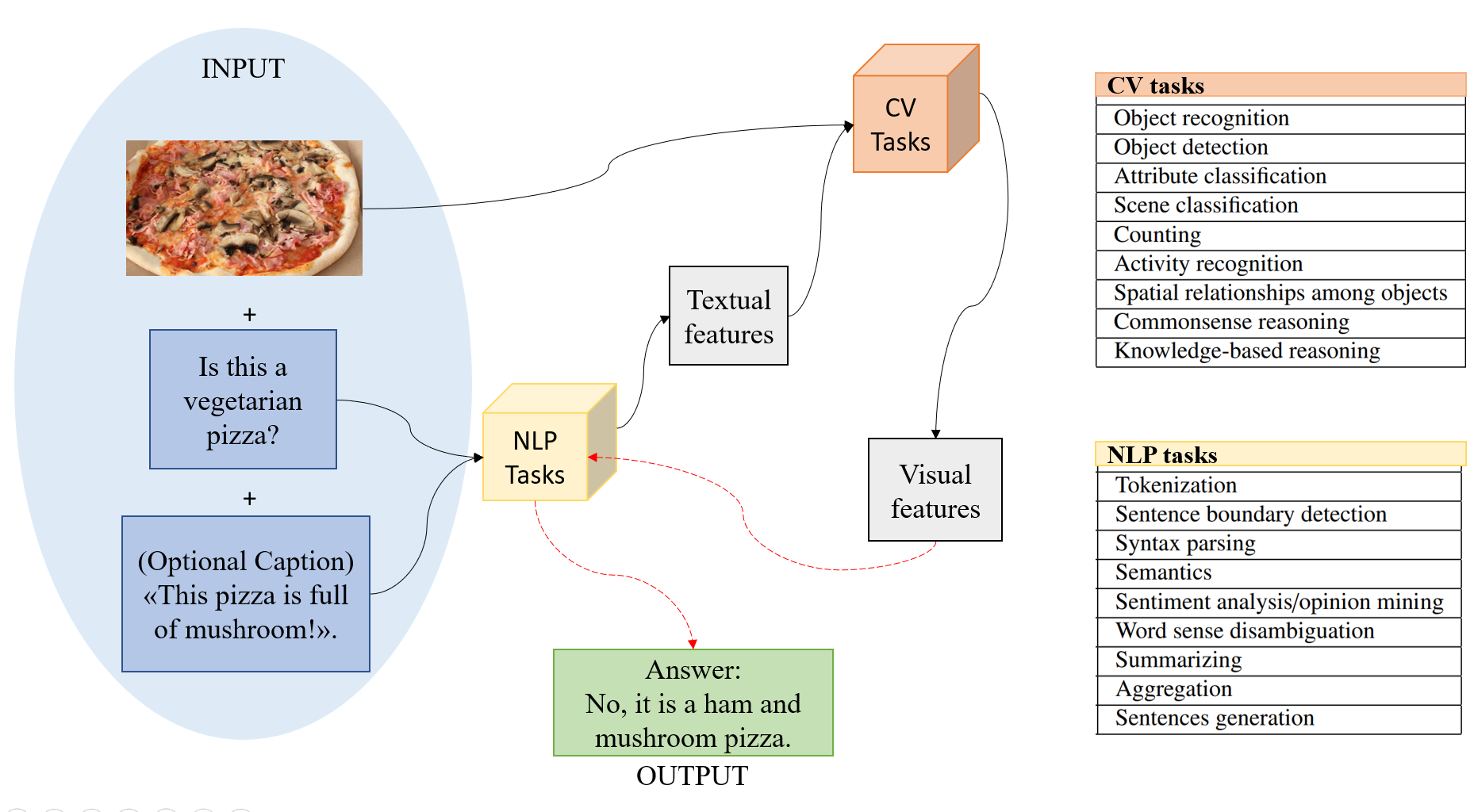}
    \caption{An example of a workflow of VQA method on the left, some of the CV tasks and NLP tasks involved in VQA are on the right.}
    \label{flow}
\end{figure*}

Visual Question Answering (VQA) is at present one of the most interesting joint applications of Artificial Intelligence (AI) to Computer Vision (CV) and Natural Language Processing (NLP). Its purpose is to achieve systems capable of answering different types of questions expressed in natural language and regarding any image. To this aim, a VQA system relies on algorithms of different nature that jointly take as input an image and a natural language question about it and generate a natural language answer as output. Humans naturally succeed in this, except for special conditions, and AI aims at reproducing this ability. The role of NLP in solving this multi-disciplinary problem is to understand the question, and of course to generate the answer according to the results obtained by CV. Text-based Q\&A is a longer studied problem in NLP. The difference with VQA is that both search and reasoning regard the content of an image. A classification of CV tasks entailed by VQA can be found in the recent survey in \cite{manmadhan2020visual} and summarized in the right part of figure \ref{flow}, where we also shown the NLP tasks involved in this case. On the left part of the figure, the overall workflow of a classical VQA framework can be summarized with the proposed example. 

VQA research crosses several AI areas, including CV, NLP and also Knowledge Representation \& Reasoning (KR), the latter being able to reason and extract semantics from processed media. Several surveys discuss VQA approaches from different points of view. Among the most recent ones, \cite{manmadhan2020visual} propose an extensive analysis and comparison, among other aspects, of different methodologies underlying the different steps of VQA including featurization, for both image and question (Phase I), and joint comprehension of image and question features to generate a correct answer (Phase II). \cite{zhang2019information} devote special attention to fusion techniques adopted in Phase II, distinguishing between fusion techniques for image QA and for video QA. Similarly, \cite{wu2017visual} classify methods by their strategy to connect the visual and textual modalities and, in particular, it examines the approach of combining convolutional and recurrent neural networks to map images and questions onto a common feature space. Looking at these surveys, it is possible to observe that their attention is focused on methodological proposals, generally neglecting the possible application domains. The latter are only shortly listed as those where the VQA can be useful. The set mentioned by \cite{manmadhan2020visual} includes: to help blind users to communicate through pictures, to attract customers of online shopping sites by giving "semantically" satisfying results for their search queries, to allow learners engaged in educational services to interact with images, to help the analysts in surveillance data analysis to summarize the available visual data. The authors also hypothesize as Visual Dialogue, envisaged as a successor of VQA,  can even be used to give natural language instructions to robots. A similar list is presented in \cite{zhang2019information}: blind person assistance (the most popular according to the citations achieved), autonomous driving, smart camera processing on food images, implementation of robot tutors with the function of automatic math problem solvers, execution of trivial tasks such as "\textit{spotting an empty picnic table at the park, or locating the restroom at the other end of the room without having to ask}." A more general use is mentioned by \cite{kafle2017visual} for advanced image retrieval. Without using image meta-data or tags, it could be possible, e.g., to find all images taken in a certain setting: one might simply ask `Is it raining?' for all images in the dataset without using image annotations. The general-purpose nature of most VQA-related literature is also reflected by the datasets exploited as benchmarks: even in the review papers, the surveyed datasets are mostly general-purpose ones, where images are possibly classified in natural, clip art or synthetic as in \cite{wu2017visual}, with no reference to a specific source domain. Only the work in \cite{kafle2017visual} focuses, among the other topics, on criticizing some current popular datasets with regard to their ability to properly train and assess VQA algorithms, and on proposing new features for future VQA benchmarks. Among the others, the authors mention larger size and also lower amount of bias, since current VQA systems are considered more dependent on the questions and how they are phrased than the image content. The additional requirement explored here is that the dataset used for performance evaluation should also be related to the VQA domain, if this presents specific conditions. As a matter of fact, most  works (relatively fewer than general-purpose ones) tackling a specific application domain also propose suitable related datasets.\\

The aim of the present paper is to survey VQA proposals from a novel point of view, and to investigate at which extent different application domains inspire different kinds of questions and call for different benchmarks and/or approaches. An extensive literature search reveals that relatively few papers have tackled specific domains. The following sections will focus on these. The most popular special application of VQA in literature is the support for automatic intelligent medical diagnosis, that deserves a large section. It encompasses different kinds of problems, characterized by different kinds of image capture technologies and image content types. The aid to blind and visually impaired individuals follows, enabling them to get information about images both on the web and in the real world, e.g., in advanced domotics. A kind of anticipation though without implementation is already envisaged in \cite{munoz2006perceptual}. It is worth noticing that these two domains have inspired non only the collection of specific ad-hoc datasets, but also their use as benchmarks in domain-related international challenges. A much lower number of devoted works deal with unattended surveillance, with proposals for systems able to relief a human operator from the burden of continuous attention and to raise an alarm in anomalous situations. Social and cultural purposes inspire systems addressing advanced education and personalized fruition of cultural heritage, and smart and customer-tailored advertisement. The paper will finally report about some very recent works focusing on the novelty of either the kind of data taken into account or of the new approaches to questioning/answering.\\

The paper proceeds as follows. Sections from \ref{sec:med} to \ref{sec:adv} present recent works in the domains of medical VQA, support for blind people, video surveillance, education and cultural heritage, and advertising; Section \ref{sec:emch} presents emerging approaches for new kinds of data and new questioning/answering strategies. Section \ref{sec:final} briefly points out some concluding remarks.


\begin{figure} [ht!]
    \centering
    \includegraphics[width=1\columnwidth]{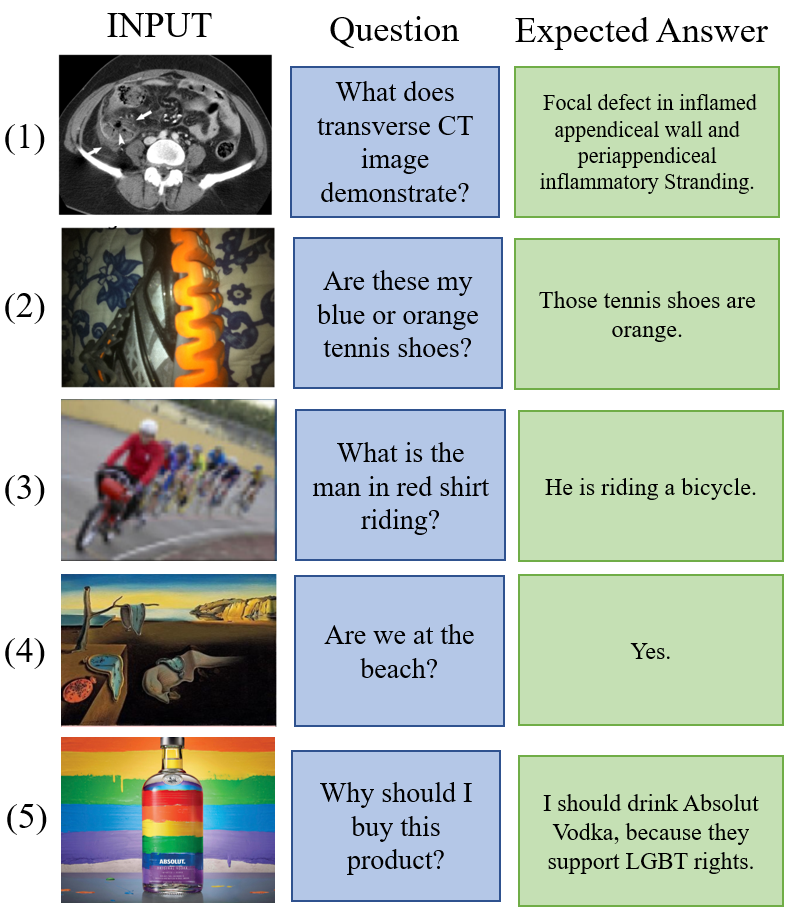}
    \caption{Examples of various VQA applications: (1) Medical image from ImageCLEF 2018 VQA-Med (\cite{abacha2019vqa}); (2) For Visually impaired people, from VizWiz Dataset (\cite{vizwiz}); (3) Video Surveillance image from BOAR Dataset (\cite{toor2019question}); (4) Image of a Painting from Artpedia (\cite{10.1007/978-3-030-30645-8_66}); (5) Advertising image from \cite{Hussain_2017_CVPR}.}
    \label{VISNIRDataset}
\end{figure}

\section{Medical VQA}
\label{sec:med}

AI-based medical image understanding and related medical questions-answering (from here on, med-VQA) is recently attracting increasing interest by researchers. In fact, this topic is opening new scenarios for supporting medical staff in taking clinical decision, as well as for enhanced diagnosis through computer-based "second opinion". However, the experimentation of any approach is conditioned by the availability of a dedicated database including medical images, of possibly specific type, and related QA pairs. These requirements have been first addressed by the ImageCLEF 2018 evaluation campaign for the Medical Domain Visual Question Answering pilot task, as described in \cite{hasan2018overview}. The related first med-VQA public dataset included a total of 2,866 medical images, 2,278 of which used for training, 264 for testing and 324 for validation, along with 6,413 QA pairs. In The following ImageCLEF 2019 edition (\cite{abacha2019vqa}), a larger dataset containing 4,200 radiology images as well as 15,992 QA pairs was released, with a wide variety of imaging modalities, type of organs and pathologies. All the works resumed in the following have based the reported experiments on one of the aforementioned datasets. More recently, the introduction of two new datasets, namely VQA-RAD presented in \cite{lau2018dataset}, and PathVQA described in \cite{he2020pathvqa}, promises to further improve the variety and specificity of training and test samples for this challenging declination of VQA. A broad range of deep frameworks has been proposed to address the requirements of med-VQA. The authors of \cite{vu2020question} propose a med-VQA deep learning approach exploiting a multimodal question-centric strategy fusing together the image and the written question in the query, assigning a greater fusion weight to the latter. The fusion mechanism combines question and image features to achieve maximum adherence to the query sentence. The answer to the query can be of different types, ranging from binary and numbers to short sentences. The achieved accuracy exceed state-of-the-art. In \cite{ren2020cgmvqa}, a novel method is presented to break the complex med-VQA problem down into multiple simpler problems through a classification and generative model. To this aim, the proposed model uses data augmentation as well as text tokenization, switching between classification and generative models by changing both output layer and loss function while retaining the core component. The generative model is built masking position by position instead of using an encoder-decoder framework. Transfer learning and multi-task learning within a modular pipeline architecture are used in \cite{kornuta2019leveraging} to cope with the wide variety of images in the ImageCLEF-2019 dataset by extracting its inherent domain knowledge. The proposed Cross Facts Network basically exploits upstream tasks to cross-utilize information useful to increase the precision on more complex downstream tasks. This results in a clear score improvement on the validation set. On a similar line of research the authors of \cite{liu2019effective} propose ETM-Trans, a deep transfer learning approach based on embedded topic modelling applied to textual questions, through which topic labels are associated to medical images for fine tuning the pre-trained ImageNet model. A co-attention mechanism is also exploited, where residual networks is used to provide fine-grained contextual information for answer derivation. In \cite{zhou2018employing} a CNN based Inception-Resnet-v2 model is used to extract image features along with a RNN based Bi-LSTM model to encode questions. The concatenation of image features and coded questions is therefore used to generate the answers. A normalization step, including both image enhancement techniques and questions lemmatization is performed beforehand. The shortage of large labeled datasets to effectively train deep learning models for med-VQA is the focus of \cite{nguyen2019overcoming}.  The authors explore the use of an unsupervised denoising auto-encoder to leverage the availability of large quantities of unlabelled medical images to achieve trained weights that can be more easily adapted to the med-VQA domain. Moreover, they also exploit supervised meta-learning to learn meta-weights which can adapt to the domain of interest requiring only a small labeled training set. The authors of \cite{lubna2019mobvqa} present a convolutional neural-network-based med-VQA system, aimed at providing answers according to input image modalities such as X-ray, computer-tomography, magnetic resonance, ultrasound, positron emission  tomography etc. where the image modality can also be identified by the system. On a similar line of research, in \cite{bghiel2019visual} a CNN is used to process medical image queries along with a RNN encoder-decoder model to encode image and question input vectors and to decode the states required to predict target answers. These are generated in natural language as output by means of greedy search algorithm. An encoder-decoder model is also at the core of \cite{allaouzi2019encoder}. Here a pre-trained CNN model is used in the encoding step along with LSTM model to embed textual data. Another deep learning inspired approach is the one proposed in \cite{allaouzi2018deep} where a combination of CNN and bi-directional Long Short Term Memory (LSTM) coupled with a decision tree classifier is used to address the med-VQA problem in terms of a multi-label classification task. According to this method, each  label is associated to a unique word among those included in the answer dictionary previously built upon the training set. Similarly, the authors of \cite{peng2018umass} exploit residual networks of a deep learning framework to extract visual features from the input image as a result of its interaction with LSTM representation of the question. The goal is to achieve small granularity context data useful to derive the answer. Efficient visual-textual feature integration is achieved through Multi-modal Factorized High-order as well as Multi-modal Factorized Bilinear pooling. LSTM and Support Vector Machine (SVM) are explored in \cite{shi2019deep}. LSTM is used for extracting textual features from questions along with image features thanks to transfer learning and co-attention mechanism. An SVM based model is trained to predict what category a question belongs to, providing an additional feature. All the resulting features are then efficiently integrated by means of a multi-modal factorized high-order pooling technique.  \cite{al2019just} propose a med-VQA model based on differently specialized sub-models, each optimized to answer to  a specific class of questions. Considered image classification sub-models include "modality", "abnormality", "organ systems" and "plane"  that are defined through pre-trained VGG16 network. Since questions related to each type are repetitive, the approach is not based on them to predict the answers, yet they are used to choose the best suited model to produce the answers and their format. A CNN based on VGG16 network is also exploited in \cite{yan2019zhejiang} along with global average pooling strategy to extract medical image features by means of a small-size training set. A BERT model is used to encode the semantics behind the input question and then a co-attention mechanism enhanced by jointly learned attention is used for feature fusion. A bilinear model aimed at grouping and synthesizing extracted image and question features for med-VQA is proposed by \cite{vu2019ensemble}. This model exploits an attention-based scheme to restrict on relevant input context, instead of relying on additional training data. Additionally, the method is also boosted by an ensemble of trained models. On a different line of research, image captioning and machine translation are explored by the authors of \cite{ambati2018sequence}, aiming at generating an answer to the image-question pair in terms of a sequence of words. Image captioning requires an accurate image understanding, and similarly machine translation requires an accurate comprehension of the input sequence to effectively translate it. This approach provided the highest accuracy scores for the ImageCLEF 2018 challenge. Stacked attention Network (SAN) along with Multimodal Compact Bilinear Pooling (MCB) VQA models are used in \cite{abacha2018nlm}. In this approach, both models rely on CNNs for image processing, respectively VGG-16 for SAN and ResNet-152 for MCB, while LSTMs are exploited for question processing. Their final hidden layer provides question vector extraction. In \cite{bansalmedical}, the proposed med-VQA pipeline partitions questions into two classes. The first one requires answers to come from a fixed pool of categories, while the second one requires to generate answers based on abnormal visual features in the input image. The first class is defined by using Universal Sentence Encoder question embeddings and ResNet image embeddings, feeding an attention-based mechanism to generate answers. The second class uses the same ResNet image embedding along with word embeddings from a Word2Vec model. This is pre-trained on PubMed data which is used as an input to a sequence to sequence model which generates descriptions of abnormalities.

\section{VQA for visually impaired people}
Assistance to blind people is among the objectives of several VQA applications proposed in the recent years. This is mainly due to the ability of automatic VQA to answer daily questions which may help visually impaired people to live without visual barriers. During the last ten years there has been a quite fast evolution which led from needing the aid from volunteers or workers paid for answering blind's questions \citep{10.1145/1866029.1866080,10.1145/2513383.2517033}, to the automatic analysis of images and related questions, to extract and generate the proper answers. To this aim, the dataset proposed and described in \cite{vizwiz} contains 31000 visual questions originated by blind people who took a photo with their mobile phone and recorded a spoken question about it. Each image is labelled with 10 crowdsourced answers. A "privacy preserving" version of the dataset is released \citep{gurari2019vizwiz}, where image private regions are removed (credit card numbers, subject information on medical prescription, etc.). An iPhone application, named VizWiz \cite{bigham2010vizwiz} allows asking a visual question and obtaining an answer in nearly real time. The authors of \cite{Anderson_2018_CVPR} propose a combined bottom-up and top-down attention mechanisms in which the question is analyzed by means of a GRU and the image is processed by a CNN. The vectors are then combined to produce a set of scores over the candidate answers. The bottom-up mechanism is based on a Faster-RCNN, which submits to the model the image regions together with the labels, while the top-down mechanism weighs the image features by applying a weighted sum with the GRU output. Another interesting application aimed at helping blind people is described in \cite{weiss2019navigation}, in which the authors exploit a reinforcement learning model in order to help a blind person to navigate the street. 

\section{VQA in Video Surveillance scenarios}\label{sec:videosurv}
The adoption of a VQA approach in video surveillance scenarios may help operators to enhance the understanding of a scene, thus helping them to take fair and faster decisions. The authors of \citep{li2019isee} propose a complete platform called ISEE for parsing large video surveillance data. The platform is organized in three modules, which are distributed on both CPU and GPU cluster: (i) detection and tracking module, (ii) attribute recognition module, and (iii) re-identification module. The first module exploits a Gaussian Mixture Model for the analysis on the CPU and a Single Shot multibox detector with a Faster R-CNN for the detection on the GPU. Both use the Nearest Neighbor-based tracker. The second module exploits DeepMAR and LSPR\_attr for the attribute recognition; the third exploits LSPR\_ReId and MSCAN for re-identification. The system has been tested over the RAP dataset \citep{8510891}. The authors of \cite{TOOR2019111}  mostly focus on the soft biometric aspects of the Q\&A in video surveillance, thus proposing C2VQA-BOARS (Context and Collaborative Visual Question Ansqering for Biometric Object-Attribute Relevance and Surveillance). The system answers a question by fusing information from the question itself with the caption obtained by an analysis of the image. Three models are proposed: (i) C2VQA-All uses a set of BiLSTMS to encode question and caption; four equally weighted training objectives are used to train the model: question relevance, type of object in the question, the attributes of the object and the final classification for the relevance of the object; (ii) C2VQA-Image takes a set of GloVe word-embeddings and uses a 2-layer LSTM for question encoding; the question is combined with the dense vector obtained by feeding a pre-trained ResNet-50 model with the considered image; (iii) C2VQA-Rel is similar to C2VQA-All, but only takes the binary relevance of the question and the final classification of the object attribute relevance. More conceptual approaches can be found in \cite{Katz2003AnsweringQA} and in \cite{6818956}. In the former, the authors propose a system for supporting the question answering operation about moving objects in videos, by filtering the trajectory information of the objects (people and vehicles), and representing the movements by means of a structured annotation. These annotations can be easily navigated for obtaining answers over movement-related questions ("\textit{Which direction is the red car going?}", "\textit{Did any cars leave the garage?}"...). The latter work, instead, proposes an ontology/graph based taxonomy schema for describing events in the video and associated captions. A probabilistic generative model is then used for capturing the relations between input video and input question.

\section{VQA Education and cultural heritage}\label{sec:educu}
One of the main aspects of VQA is its high correlation to human perception. 
Even if a VQA system can focus the attention on different parts of an image compared to humans, it is proven that devising a VQA architecture "interested" in the same image parts is possible \citep{DAS201790}. 
The inverse process can be also be carried out. 
\cite{8291426} developed and tested an educational robot using VQA to formulate questions and start an educational dialog taking inspiration from the surrounding environment, using a faster R-CNN. This system shows a great ability to improve the children's desire to explore. VQA can improve such desire for adults too, in particular for cultural heritage. \cite{Bongini2020VisualQA} propose to explore museums and art galleries using VQA to interact with an audio-guide. The authors use many classical VQA datasets and a cultural dataset named Artpedia \citep{10.1007/978-3-030-30645-8_66} to feed two BERT modules, for question classification and answering, and a faster R-CNN for the VQA module. They annotated 30 images of Artpedia with 3 or 5 Q\&A to perform tests. As a result, the user can directly ask  questions he/she is interested in, avoiding long descriptions and freely navigating through the elements of the painting or the sculpture. This way to explore art can replace static audio-guides and the growing interest in this goal has inspired to the construction of a dedicated dataset \citep{Sheng2016ADF}. This dataset is focused on the old-Egiptyan Amarna period and contains 16 artworks, 805 questions, 204 documents, 101 related documents and 139 related paragraphs in English. This dataset is very specific and quite limited, but it is an interesting starting point to apply VQA in the service of art.

\section{VQA and Advertising}\label{sec:adv}
Advertising is something strongly related to image understanding. A user looking at an advertise not only sees the objects inside the scene, but also the related text and the relations among the objects, and interprets all such information within a precise cultural context. An advertise must be quite simple to be understandable for the greatest number of people and at the same time interesting and eye-catching.
No surprise than that VQA can find a challenging field of application in advertising. 
The first task to complete is using VQA to understand the advertise and, in particular, the underlying communicative strategy. \cite{Hussain_2017_CVPR} present two datasets for this purpose, one with images and one with videos. The image dataset contain 64832 ads. For each advertise there is a set of Q\&A about what the client is led to do by it, for a total of 202090 elements with 38 topics, 30 sentiments and 221 symbols. The video dataset has 3477 elements, with 3 or 5 questions per video and the same symbols, sentiments and topics of images. The authors use a two-layer LSTM and VGGNet to decode images ads and 152-layer ResNets to decode video ads.
Once an automatic system is able to understand the meaning of an advertise, it is natural to ask whether it is possible to automatically choose which ads to show. 
The authors of \cite{park2019ads} focus their research on predicting the users' preferences by understanding what impress them most. They built Real-ad, a dataset of 3747 images with 40 attributes and collected about 500 millions of impressions from the users. They include VQA in a low level fusion using LSTM, followed by an attention mechanism and a high level fusion to emphasize the relations between visual and auxiliary information. The result is an attention heatmap that shows the parts of the images from which the potential client is attracted.
Finally, the most challenging step will be to automatically find the most successful ads to help advertising designers. 
\cite{zhou2020recommending} propose a way to use VQA to extract relevant information about past campaigns in multi-source data as texts and images. The authors use 64000 images from the dataset in \cite{Hussain_2017_CVPR} and extract information, also searching online, analyse keyphrases and generate new possible advertises. To this aim, they use a cross-modality encoder architecture followed by a feed forward network. Even if not exploded yet, this task shows an interesting research directions.

\begin{table*}[t]
\centering
\scriptsize
\caption{A summary of application domains tackled by VQA literature with a summary of the features of the datasets used as benchmarks. Also, the approaches are reported, with the best results for each dataset. Following a legend for interpret the table: 
Dataset: C:Classes, L:Labels.
Dataset Size: AS: Audio Scenes, C:Classes, T:Text, HoV:Hours of Video, I:images, Ic:Icons, V:Videos.
Notes: A:Answers, At:Attributes, C:Classes, Cap:Caption, I:Images, Imp:Impression, Q:Questions. 
Approaches: AM:Attention Mechanism, BM:Bert Model, CM:Classifier Model, DN:DenseNet, ED:Encoder-Decoder, F:Fusion, FRCNN: Faster-RCNN, GM:Generative Model, IA: Image Attention, IN:ImageNet, IV2: InceptionV2, MLA:MultiLevel Attention, MMFN:Multi-Step Modality Fusion Network, QCM:Question Classifier Model, RM:Reasoning Module, RN:ResNet, TE:Text Embedding, WE:Word Embedding,W2V:Word2Vec.
Best Results: ACC:Accuracy, BLEU:Bilingual Evaluation Understudy Score: TDA:Task Dendendant Accuracy, WBSS:Word-Based Semantic Similarity}
\label{t:final}
\begin{tabular}{|l|l|l|l|l|l|l|}
\hline
Domain & Reference(s) & Dataset & Dataset Size & Notes & Approach & Best Result \\ \hline
 & \cite{vu2020question} & ImageCLEF 2019 & 3200 I & 12,792 Q\&A & RN152+BM (AM+F) & \textbf{\cite{ren2020cgmvqa}} \\
 & \cite{ren2020cgmvqa} &  &  &  & RN152+TE (GM) & ACC: 64\% \\
 & \cite{allaouzi2019encoder} &  &  &  & DN121+LSTM (ED) & BLEU: 65,9\% \\
 & \cite{bansalmedical} &  &  &  & VQACM (Seq-to-Seq) &  \\
 & \cite{al2019just} &  &  &  & QCM+VGG &  \\
 & \cite{bghiel2019visual} &  &  &  & RN50+W2V (ED) &  \\
 & \cite{lubna2019mobvqa} &  &  &  & IN+NLP &  \\
 & \cite{vu2019ensemble} &  &  &  & RN152+BM (AM) &  \\
Medical & \cite{yan2019zhejiang} &  &  &  & VGG16+Enc (Co-AM) &  \\
 & \cite{shi2019deep} &  &  &  & RN152+LSTM (Co-AM+MFH) &  \\
 & \cite{kornuta2019leveraging} &  &  &  & ED+RM &  \\ \cline{2-7} 
 & \cite{zhou2018employing}, & ImageCLEF 2018 & 2866 I & 6,413 Q\&A & IV2+BiLSTM (AM) & \cite{ambati2018sequence} \\
 & \cite{allaouzi2018deep} &  &  &  & VGG16+BDLSTM & BLEU: 0,188 \\
 & \cite{ambati2018sequence} &  &  &  & VGG16+GRU (ED) & WBSS: 0,209 \\
 & \cite{abacha2018nlm} &  &  &  & VGG16+LSTM (SANet &  \\
 & \cite{peng2018umass} &  &  &  & RN+LSTM (Co-AM+MFP F) &  \\
 & \cite{liu2019effective} &  &  &  & IN+LSTM (Co-AM) &  \\ \cline{2-7} 
 & \cite{nguyen2019overcoming} & VQA-RAD & 315 I & 3515 Q\&A & CNNEnc.+WE (Co-AM) & ACC: 74,1\% \\ \hline
 & \cite{gurari2019vizwiz} & VizWiz-Priv & 5537 I & 1403 Q\&A & Several Approaches & - \\
Visually & \cite{vizwiz} & VizWiz & 31K I & \begin{tabular}[c]{@{}l@{}}31000 Q\\ 10 A each\end{tabular} & Several Approaches & - \\
impaired & \cite{Anderson_2018_CVPR} & VQAv2.0 & 1.1M I & 1.1M Q\&A & FRCNN+RN101+LSTM & ACC: 70,34\% \\
 & \cite{weiss2019navigation} & SEVN Simulator & - & - & RL-based Approach & ACC: 74,8\% \\ \hline
Video & \cite{li2019isee} & RAP & 587 HoV & \begin{tabular}[c]{@{}l@{}}84928 I\\ 72 At\end{tabular} & - & - \\
Surveillance & \cite{TOOR2019111} & BOAR + BTV & \begin{tabular}[c]{@{}l@{}}$\sim$45K I\\ 5 V\end{tabular} & \begin{tabular}[c]{@{}l@{}}$\sim$23K Q\\ 101 C\end{tabular} & RN50+BiLSTM & ACC: 61,36\% \\
 & \cite{6818956} & Homemade & 2 V & 5 Cap each & - & - \\ \hline
Cultural & \cite{Bongini2020VisualQA} & Annotated Artpedia & 30 I & $\sim$120 Q\&A & FRCNN+BM & ACC:25,1\% \\
Heritage & \cite{Sheng2016ADF} & Homemade & 16 I, 444 T & 805 Q\&A & - & - \\ \hline
Advertising & \cite{Hussain_2017_CVPR} & Homemade & \begin{tabular}[c]{@{}l@{}}64832 I, \\ 3477 V\end{tabular} & $\sim$273k Q\&A & RN+LSTM & TDA \\
 & \cite{park2019ads} & Homemade & 3747 I & 500k Imp & MMFN & - \\ \hline
 & \cite{Chou2020VisualQA} & Homemade & 1490 I in $360^\circ$ & 16945 Q\&A & Tucker\&Diffusion (MLA) & ACC: 58,66\% \\
 & \cite{sampat2020diverse} & Homemade & \begin{tabular}[c]{@{}l@{}}10209 I \\ 9156 T\end{tabular} & 9267 Q\&A & CNN enc.(I\&Q) & ACC: 39,63\% \\
 & \cite{wang2020long} & TACoS-QA & 185 V & 21310 Q\&A & 3DCNN+LSTM & ACC: 24,82\% \\
 & \cite{wang2020long} & MSR-VTT-QA & 3852 V & 19748 Q\&A &  &  \\
 Misc & \cite{abdelnour2019visual} & Homemade & 7500 AS & 300000 Q\&A & \begin{tabular}[c]{@{}l@{}}FiLM Network\\ MAC Network\end{tabular} & \begin{tabular}[c]{@{}l@{}}ACC: 90,3\%\\ ACC: 44,8\%\end{tabular} \\
 & \cite{hosseinabad2020multiple} & Homemade & 85321 Ic & 429654 Q\&A & CNN+LSTM & ACC: 25,78\% \\
 & \cite{bhattacharya2019does} & VQA 2.0+VizWIz+L & 44955 I & 224775 Q\&A & CNN+GRU & ACC: 44,55\% \\
 & \cite{toor2019question} & COCO-A+VG+C & 19431 I, 342 C & $\sim$80K Q\&A & GloVe+BiLSTM & ACC: 79,32\% \\
 & \cite{teney2019actively} & VQA-CPv2+COCO & 325721 I & \begin{tabular}[c]{@{}l@{}}$\sim$1.5M Q\\ $\sim$15M A\end{tabular} & IA+WE & ACC: 34,25\% \\ \hline
\end{tabular}
\end{table*}

\section{Emerging challenges/Misc}\label{sec:emch}
Emerging lines in VQA mainly focus on new input or Q\&A. \\
Concerning the data, \cite{Chou2020VisualQA} explore the use of $360^\circ$ images on which the information can be located in all the field of view. They used 1490 indoor images from two 3D datasets and generated 16945 Q\&A. A cubemap-based architecture extracts visual features and, then, a multi-level attention network aggregates features. Other than image, a VQA system may be required to extract information also from text. \cite{sampat2020diverse} built a dataset to test modalities in which an image-text joint inference is required. The dataset is composed by educational, web and other VQA datasets resources, for a total of 10209 images, 9156 different texts and 9267 questions. They also tested this new dataset using existing VQA models discovering that the latter do not fit well those new data. 
The video-question-answering is also a well-known field, however the long-video QA is unexplored. \cite{wang2020long} manage this kind of data in their work building two datasets of long video, TACoS-QA and MSR-VTT-QA, containing 187 and 3852 videos, respectively. The number of Q\&A is about 20000 per dataset. Based on that, they develop a matching-guided attention model for video and question embedding, question-related video content localization
and answer prediction.
The VQA know-how is extending to Acustic Question Answering (AQA). \cite{abdelnour2019visual} show how to create auditory scenes and related Q\&A. Their 7500 scenes and 300000 Q\&A were tested using two neural networks: FiLM, based on Conditional Batch Normalization, and MAC, based on LSTM models. \\
Concerning the new challenges that can emerge for Q\&As, the more intuitive is the possibility of multiple answers. Both \cite{hosseinabad2020multiple} and \cite{bhattacharya2019does} explore this possibility. The former start this research building a dataset of 100 simple icons randomly located in 85321 images for a total of 429654 Q\&A. They tested this new dataset building a LSTM-based neural network. 
The latter focus on the reason why a question can have more than one answer. They built a dataset, starting from two popular datasets, VQA 2.0 \citep{10.1109/ICCV.2015.279} and VizWiz  \citep{vizwiz}, for a total of 44955 images and 224775 annotations, labelling 9 reason for different answers. They built a prediction answer network based on CNN and attention model in which not only the answer is predicted but also a probable reason for difference.
Studying the properties of the questions is gaining popularity. \cite{toor2019question} not only built a system to detect when a question is not related to the image but also a method to edit this question. Starting from the COCO-A \citep{BMVC2015_52} and the VG \citep{krishna} datasets, they obtained 19431 images and 55738 Q\&A for the question relevance and 22172 questions about 342 classes for the question editing. Their method is then based on BiLSTM embedded in a neural network architecture.
From all those works it is clear that a huge amount of Q\&A is required to built a well generalising system, however collecting them is time-consuming. For this reason, \cite{teney2019actively} propose a new VQA system in which the ability to answer an unknown question is obtained fusing known questions and external data. For this purpose they focus on weight adaptation on a basic VQA model, using VQA-CP v2 dataset \citep{Agrawal2018DontJA} and COCO captioning dataset. \\
Finally, to evaluate the effectiveness of a VQA system, \cite{8528867} propose an Inverse VQA. They use a question encoder that encodes the image and question, and a decoder that, from the features of the image, generates a visual question. The question encoder is based on an LSTM architecture and the authors show how to use this method to perform VQA diagnosis.

\section{Discussion and conclusions}\label{sec:final}
Two elements are interesting to underline. First, in most cases, even though specific datasets are collected, methods are mostly inherited though sometimes adapted. In general, except for med-VQA, there is no attempt for boost optimization following a stronger domain characterization, except for new kinds of data. It can be hypothesized that when the image features are very different than usual, as in medical imaging, this could take to a more effective design of the entailed processing. A second consideration concerns the size of the datasets. While this problems is underlined in the general case in \cite{kafle2017visual}, ad-hoc datasets seldom reach the huge amount of information requested for a robust generalizability of the obtained performance. These two aspects definitely deserve attention from the VQA community in order to finally reach performance able to boost real world domain-specific applications.

\bibliographystyle{model2-names}
\bibliography{VQA_bib}

\begin{thebibliography}{60}
\expandafter\ifx\csname natexlab\endcsname\relax\def\natexlab#1{#1}\fi
\providecommand{\url}[1]{\texttt{#1}}
\providecommand{\href}[2]{#2}
\providecommand{\path}[1]{#1}
\providecommand{\DOIprefix}{doi:}
\providecommand{\ArXivprefix}{arXiv:}
\providecommand{\URLprefix}{URL: }
\providecommand{\Pubmedprefix}{pmid:}
\providecommand{\doi}[1]{\href{http://dx.doi.org/#1}{\path{#1}}}
\providecommand{\Pubmed}[1]{\href{pmid:#1}{\path{#1}}}
\providecommand{\bibinfo}[2]{#2}
\ifx\xfnm\relax \def\xfnm[#1]{\unskip,\space#1}\fi
\bibitem[{Abacha et~al.(2018)Abacha, Gayen, Lau, Rajaraman and
  Demner-Fushman}]{abacha2018nlm}
\bibinfo{author}{Abacha, A.B.}, \bibinfo{author}{Gayen, S.},
  \bibinfo{author}{Lau, J.J.}, \bibinfo{author}{Rajaraman, S.},
  \bibinfo{author}{Demner-Fushman, D.}, \bibinfo{year}{2018}.
\newblock \bibinfo{title}{Nlm at imageclef 2018 visual question answering in
  the medical domain.}, in: \bibinfo{booktitle}{CLEF (Working Notes)}.
\bibitem[{Abacha et~al.(2019)Abacha, Hasan, Datla, Liu, Demner-Fushman and
  M{\"u}ller}]{abacha2019vqa}
\bibinfo{author}{Abacha, A.B.}, \bibinfo{author}{Hasan, S.A.},
  \bibinfo{author}{Datla, V.V.}, \bibinfo{author}{Liu, J.},
  \bibinfo{author}{Demner-Fushman, D.}, \bibinfo{author}{M{\"u}ller, H.},
  \bibinfo{year}{2019}.
\newblock \bibinfo{title}{Vqa-med: Overview of the medical visual question
  answering task at imageclef 2019}, in: \bibinfo{booktitle}{CLEF2019 Working
  Notes. CEUR Workshop Proceedings}, pp. \bibinfo{pages}{09--12}.
\bibitem[{Abdelnour et~al.(2019)Abdelnour, Salvi and
  Rouat}]{abdelnour2019visual}
\bibinfo{author}{Abdelnour, J.}, \bibinfo{author}{Salvi, G.},
  \bibinfo{author}{Rouat, J.}, \bibinfo{year}{2019}.
\newblock \bibinfo{title}{From visual to acoustic question answering}.
\newblock \bibinfo{journal}{arXiv preprint arXiv:1902.11280} .
\bibitem[{Agrawal et~al.(2018)Agrawal, Batra, Parikh and
  Kembhavi}]{Agrawal2018DontJA}
\bibinfo{author}{Agrawal, A.}, \bibinfo{author}{Batra, D.},
  \bibinfo{author}{Parikh, D.}, \bibinfo{author}{Kembhavi, A.},
  \bibinfo{year}{2018}.
\newblock \bibinfo{title}{Don't just assume; look and answer: Overcoming priors
  for visual question answering}.
\newblock \bibinfo{journal}{2018 IEEE/CVF Conference on Computer Vision and
  Pattern Recognition} , \bibinfo{pages}{4971--4980}.
\bibitem[{Al-Sadi et~al.(2019)Al-Sadi, Talafha, Al-Ayyoub, Jararweh and
  Costen}]{al2019just}
\bibinfo{author}{Al-Sadi, A.}, \bibinfo{author}{Talafha, B.},
  \bibinfo{author}{Al-Ayyoub, M.}, \bibinfo{author}{Jararweh, Y.},
  \bibinfo{author}{Costen, F.}, \bibinfo{year}{2019}.
\newblock \bibinfo{title}{Just at imageclef 2019 visual question answering in
  the medical domain}.
\newblock \bibinfo{journal}{Working Notes of CLEF} .
\bibitem[{Allaouzi and Ahmed(2018)}]{allaouzi2018deep}
\bibinfo{author}{Allaouzi, I.}, \bibinfo{author}{Ahmed, M.B.},
  \bibinfo{year}{2018}.
\newblock \bibinfo{title}{Deep neural networks and decision tree classifier for
  visual question answering in the medical domain.}, in:
  \bibinfo{booktitle}{CLEF (Working Notes)}.
\bibitem[{Allaouzi et~al.(2019)Allaouzi, Benamrou and
  Ahmed}]{allaouzi2019encoder}
\bibinfo{author}{Allaouzi, I.}, \bibinfo{author}{Benamrou, B.},
  \bibinfo{author}{Ahmed, M.}, \bibinfo{year}{2019}.
\newblock \bibinfo{title}{An encoder-decoder model for visual question
  answering in the medical domain}.
\newblock \bibinfo{journal}{Working Notes of CLEF} .
\bibitem[{Ambati and Dudyala(2018)}]{ambati2018sequence}
\bibinfo{author}{Ambati, R.}, \bibinfo{author}{Dudyala, C.R.},
  \bibinfo{year}{2018}.
\newblock \bibinfo{title}{A sequence-to-sequence model approach for imageclef
  2018 medical domain visual question answering}, in: \bibinfo{booktitle}{2018
  15th IEEE India Council International Conference (INDICON)},
  \bibinfo{organization}{IEEE}. pp. \bibinfo{pages}{1--6}.
\bibitem[{Anderson et~al.(2018)Anderson, He, Buehler, Teney, Johnson, Gould and
  Zhang}]{Anderson_2018_CVPR}
\bibinfo{author}{Anderson, P.}, \bibinfo{author}{He, X.},
  \bibinfo{author}{Buehler, C.}, \bibinfo{author}{Teney, D.},
  \bibinfo{author}{Johnson, M.}, \bibinfo{author}{Gould, S.},
  \bibinfo{author}{Zhang, L.}, \bibinfo{year}{2018}.
\newblock \bibinfo{title}{Bottom-up and top-down attention for image captioning
  and visual question answering}, in: \bibinfo{booktitle}{The IEEE Conference
  on Computer Vision and Pattern Recognition (CVPR)}.
\bibitem[{Antol et~al.(2015)Antol, Agrawal, Lu, Mitchell, Batra, Zitnick and
  Parikh}]{10.1109/ICCV.2015.279}
\bibinfo{author}{Antol, S.}, \bibinfo{author}{Agrawal, A.},
  \bibinfo{author}{Lu, J.}, \bibinfo{author}{Mitchell, M.},
  \bibinfo{author}{Batra, D.}, \bibinfo{author}{Zitnick, C.L.},
  \bibinfo{author}{Parikh, D.}, \bibinfo{year}{2015}.
\newblock \bibinfo{title}{Vqa: Visual question answering}, in:
  \bibinfo{booktitle}{Proceedings of the 2015 IEEE International Conference on
  Computer Vision (ICCV)}, \bibinfo{publisher}{IEEE Computer Society},
  \bibinfo{address}{USA}. p. \bibinfo{pages}{2425–2433}.
\newblock \URLprefix \url{https://doi.org/10.1109/ICCV.2015.279},
  \DOIprefix\doi{10.1109/ICCV.2015.279}.
\bibitem[{Bansal et~al.(2019)Bansal, Gadgil, Shah and Verma}]{bansalmedical}
\bibinfo{author}{Bansal, M.}, \bibinfo{author}{Gadgil, T.},
  \bibinfo{author}{Shah, R.}, \bibinfo{author}{Verma, P.},
  \bibinfo{year}{2019}.
\newblock \bibinfo{title}{Medical visual question answering at image clef
  2019-vqa med.}, in: \bibinfo{booktitle}{CLEF (Working Notes)}.
\bibitem[{Bghiel et~al.(2019)Bghiel, Dahdouh, Allaouzi, Ahmed and
  Boudhir}]{bghiel2019visual}
\bibinfo{author}{Bghiel, A.}, \bibinfo{author}{Dahdouh, Y.},
  \bibinfo{author}{Allaouzi, I.}, \bibinfo{author}{Ahmed, M.B.},
  \bibinfo{author}{Boudhir, A.A.}, \bibinfo{year}{2019}.
\newblock \bibinfo{title}{Visual question answering system for identifying
  medical images attributes}, in: \bibinfo{booktitle}{The Proceedings of the
  Third International Conference on Smart City Applications},
  \bibinfo{organization}{Springer}. pp. \bibinfo{pages}{483--492}.
\bibitem[{Bhattacharya et~al.(2019)Bhattacharya, Li and
  Gurari}]{bhattacharya2019does}
\bibinfo{author}{Bhattacharya, N.}, \bibinfo{author}{Li, Q.},
  \bibinfo{author}{Gurari, D.}, \bibinfo{year}{2019}.
\newblock \bibinfo{title}{Why does a visual question have different answers?},
  in: \bibinfo{booktitle}{Proceedings of the IEEE International Conference on
  Computer Vision}, pp. \bibinfo{pages}{4271--4280}.
\bibitem[{Bigham et~al.(2010a)Bigham, Jayant, Ji, Little, Miller, Miller,
  Miller, Tatarowicz, White, White and Yeh}]{10.1145/1866029.1866080}
\bibinfo{author}{Bigham, J.P.}, \bibinfo{author}{Jayant, C.},
  \bibinfo{author}{Ji, H.}, \bibinfo{author}{Little, G.},
  \bibinfo{author}{Miller, A.}, \bibinfo{author}{Miller, R.C.},
  \bibinfo{author}{Miller, R.}, \bibinfo{author}{Tatarowicz, A.},
  \bibinfo{author}{White, B.}, \bibinfo{author}{White, S.},
  \bibinfo{author}{Yeh, T.}, \bibinfo{year}{2010}a.
\newblock \bibinfo{title}{Vizwiz: Nearly real-time answers to visual
  questions}, in: \bibinfo{booktitle}{Proceedings of the 23nd Annual ACM
  Symposium on User Interface Software and Technology},
  \bibinfo{publisher}{Association for Computing Machinery},
  \bibinfo{address}{New York, NY, USA}. p. \bibinfo{pages}{333–342}.
\bibitem[{Bigham et~al.(2010b)Bigham, Jayant, Ji, Little, Miller, Miller,
  Miller, Tatarowicz, White, White et~al.}]{bigham2010vizwiz}
\bibinfo{author}{Bigham, J.P.}, \bibinfo{author}{Jayant, C.},
  \bibinfo{author}{Ji, H.}, \bibinfo{author}{Little, G.},
  \bibinfo{author}{Miller, A.}, \bibinfo{author}{Miller, R.C.},
  \bibinfo{author}{Miller, R.}, \bibinfo{author}{Tatarowicz, A.},
  \bibinfo{author}{White, B.}, \bibinfo{author}{White, S.}, et~al.,
  \bibinfo{year}{2010}b.
\newblock \bibinfo{title}{Vizwiz: nearly real-time answers to visual
  questions}, in: \bibinfo{booktitle}{Proceedings of the 23nd annual ACM
  symposium on User interface software and technology}, pp.
  \bibinfo{pages}{333--342}.
\bibitem[{Bongini et~al.(2020)Bongini, Becattini, Bagdanov and
  Bimbo}]{Bongini2020VisualQA}
\bibinfo{author}{Bongini, P.}, \bibinfo{author}{Becattini, F.},
  \bibinfo{author}{Bagdanov, A.D.}, \bibinfo{author}{Bimbo, A.D.},
  \bibinfo{year}{2020}.
\newblock \bibinfo{title}{Visual question answering for cultural heritage}.
\newblock \bibinfo{journal}{ArXiv} \bibinfo{volume}{abs/2003.09853}.
\bibitem[{Chou et~al.(2020)Chou, Chao, Lai, Sun and Yang}]{Chou2020VisualQA}
\bibinfo{author}{Chou, S.H.}, \bibinfo{author}{Chao, W.L.},
  \bibinfo{author}{Lai, W.S.}, \bibinfo{author}{Sun, M.},
  \bibinfo{author}{Yang, M.H.}, \bibinfo{year}{2020}.
\newblock \bibinfo{title}{Visual question answering on $360^\circ$ images}.
\newblock \bibinfo{journal}{ArXiv} \bibinfo{volume}{abs/2001.03339}.
\bibitem[{Das et~al.(2017)Das, Agrawal, Zitnick, Parikh and Batra}]{DAS201790}
\bibinfo{author}{Das, A.}, \bibinfo{author}{Agrawal, H.},
  \bibinfo{author}{Zitnick, L.}, \bibinfo{author}{Parikh, D.},
  \bibinfo{author}{Batra, D.}, \bibinfo{year}{2017}.
\newblock \bibinfo{title}{Human attention in visual question answering: Do
  humans and deep networks look at the same regions?}
\newblock \bibinfo{journal}{Computer Vision and Image Understanding}
  \bibinfo{volume}{163}, \bibinfo{pages}{90 -- 100}.
\newblock \bibinfo{note}{Language in Vision}.
\bibitem[{Gurari et~al.(2019)Gurari, Li, Lin, Zhao, Guo, Stangl and
  Bigham}]{gurari2019vizwiz}
\bibinfo{author}{Gurari, D.}, \bibinfo{author}{Li, Q.}, \bibinfo{author}{Lin,
  C.}, \bibinfo{author}{Zhao, Y.}, \bibinfo{author}{Guo, A.},
  \bibinfo{author}{Stangl, A.}, \bibinfo{author}{Bigham, J.P.},
  \bibinfo{year}{2019}.
\newblock \bibinfo{title}{Vizwiz-priv: A dataset for recognizing the presence
  and purpose of private visual information in images taken by blind people},
  in: \bibinfo{booktitle}{Proceedings of the IEEE Conference on Computer Vision
  and Pattern Recognition}, pp. \bibinfo{pages}{939--948}.
\bibitem[{Gurari et~al.(2018)Gurari, Li, Stangl, Guo, Lin, Grauman, Luo and
  Bigham}]{vizwiz}
\bibinfo{author}{Gurari, D.}, \bibinfo{author}{Li, Q.},
  \bibinfo{author}{Stangl, A.}, \bibinfo{author}{Guo, A.},
  \bibinfo{author}{Lin, C.}, \bibinfo{author}{Grauman, K.},
  \bibinfo{author}{Luo, J.}, \bibinfo{author}{Bigham, J.},
  \bibinfo{year}{2018}.
\newblock \bibinfo{title}{Vizwiz grand challenge: Answering visual questions
  from blind people}, in: \bibinfo{booktitle}{Proceedings of the IEEE
  Conference on Computer Vision and Pattern Recognition}, pp.
  \bibinfo{pages}{3608--3617}.
\bibitem[{Hasan et~al.(2018)Hasan, Ling, Farri, Liu, M{\"u}ller and
  Lungren}]{hasan2018overview}
\bibinfo{author}{Hasan, S.A.}, \bibinfo{author}{Ling, Y.},
  \bibinfo{author}{Farri, O.}, \bibinfo{author}{Liu, J.},
  \bibinfo{author}{M{\"u}ller, H.}, \bibinfo{author}{Lungren, M.},
  \bibinfo{year}{2018}.
\newblock \bibinfo{title}{Overview of imageclef 2018 medical domain visual
  question answering task.}, in: \bibinfo{booktitle}{CLEF (Working Notes)}.
\bibitem[{{He} et~al.(2017){He}, {Xia}, {Yu}, {Jian}, {Meng} and
  {Chen}}]{8291426}
\bibinfo{author}{{He}, B.}, \bibinfo{author}{{Xia}, M.}, \bibinfo{author}{{Yu},
  X.}, \bibinfo{author}{{Jian}, P.}, \bibinfo{author}{{Meng}, H.},
  \bibinfo{author}{{Chen}, Z.}, \bibinfo{year}{2017}.
\newblock \bibinfo{title}{An educational robot system of visual question
  answering for preschoolers}, in: \bibinfo{booktitle}{2017 2nd International
  Conference on Robotics and Automation Engineering (ICRAE)}, pp.
  \bibinfo{pages}{441--445}.
\bibitem[{He et~al.(2020)He, Zhang, Mou, Xing and Xie}]{he2020pathvqa}
\bibinfo{author}{He, X.}, \bibinfo{author}{Zhang, Y.}, \bibinfo{author}{Mou,
  L.}, \bibinfo{author}{Xing, E.}, \bibinfo{author}{Xie, P.},
  \bibinfo{year}{2020}.
\newblock \bibinfo{title}{Pathvqa: 30000+ questions for medical visual question
  answering}.
\newblock \bibinfo{journal}{arXiv preprint arXiv:2003.10286} .
\bibitem[{Hosseinabad et~al.(2020)Hosseinabad, Safayani and
  Mirzaei}]{hosseinabad2020multiple}
\bibinfo{author}{Hosseinabad, S.H.}, \bibinfo{author}{Safayani, M.},
  \bibinfo{author}{Mirzaei, A.}, \bibinfo{year}{2020}.
\newblock \bibinfo{title}{Multiple answers to a question: a new approach for
  visual question answering}.
\newblock \bibinfo{journal}{The Visual Computer} , \bibinfo{pages}{1--13}.
\bibitem[{Hussain et~al.(2017)Hussain, Zhang, Zhang, Ye, Thomas, Agha, Ong and
  Kovashka}]{Hussain_2017_CVPR}
\bibinfo{author}{Hussain, Z.}, \bibinfo{author}{Zhang, M.},
  \bibinfo{author}{Zhang, X.}, \bibinfo{author}{Ye, K.},
  \bibinfo{author}{Thomas, C.}, \bibinfo{author}{Agha, Z.},
  \bibinfo{author}{Ong, N.}, \bibinfo{author}{Kovashka, A.},
  \bibinfo{year}{2017}.
\newblock \bibinfo{title}{Automatic understanding of image and video
  advertisements}, in: \bibinfo{booktitle}{The IEEE Conference on Computer
  Vision and Pattern Recognition (CVPR)}.
\bibitem[{Kafle and Kanan(2017)}]{kafle2017visual}
\bibinfo{author}{Kafle, K.}, \bibinfo{author}{Kanan, C.}, \bibinfo{year}{2017}.
\newblock \bibinfo{title}{Visual question answering: Datasets, algorithms, and
  future challenges}.
\newblock \bibinfo{journal}{Computer Vision and Image Understanding}
  \bibinfo{volume}{163}, \bibinfo{pages}{3--20}.
\bibitem[{Katz et~al.(2003)Katz, Lin, Stauffer and
  Grimson}]{Katz2003AnsweringQA}
\bibinfo{author}{Katz, B.}, \bibinfo{author}{Lin, J.J.},
  \bibinfo{author}{Stauffer, C.}, \bibinfo{author}{Grimson, W.E.L.},
  \bibinfo{year}{2003}.
\newblock \bibinfo{title}{Answering questions about moving objects in
  surveillance videos}, in: \bibinfo{booktitle}{New Directions in Question
  Answering}.
\bibitem[{Kornuta et~al.(2019)Kornuta, Rajan, Shivade, Asseman and
  Ozcan}]{kornuta2019leveraging}
\bibinfo{author}{Kornuta, T.}, \bibinfo{author}{Rajan, D.},
  \bibinfo{author}{Shivade, C.}, \bibinfo{author}{Asseman, A.},
  \bibinfo{author}{Ozcan, A.S.}, \bibinfo{year}{2019}.
\newblock \bibinfo{title}{Leveraging medical visual question answering with
  supporting facts}.
\newblock \bibinfo{journal}{arXiv preprint arXiv:1905.12008} .
\bibitem[{Krishna et~al.(2016)Krishna, Zhu, Groth, Johnson, Hata, Kravitz,
  Chen, Kalantidis, Li, Shamma, Bernstein and Li}]{krishna}
\bibinfo{author}{Krishna, R.}, \bibinfo{author}{Zhu, Y.},
  \bibinfo{author}{Groth, O.}, \bibinfo{author}{Johnson, J.},
  \bibinfo{author}{Hata, K.}, \bibinfo{author}{Kravitz, J.},
  \bibinfo{author}{Chen, S.}, \bibinfo{author}{Kalantidis, Y.},
  \bibinfo{author}{Li, L.J.}, \bibinfo{author}{Shamma, D.},
  \bibinfo{author}{Bernstein, M.}, \bibinfo{author}{Li, F.F.},
  \bibinfo{year}{2016}.
\newblock \bibinfo{title}{Visual genome: Connecting language and vision using
  crowdsourced dense image annotations}.
\newblock \bibinfo{journal}{International Journal of Computer Vision}
  \bibinfo{volume}{123}.
\bibitem[{Lasecki et~al.(2013)Lasecki, Thiha, Zhong, Brady and
  Bigham}]{10.1145/2513383.2517033}
\bibinfo{author}{Lasecki, W.S.}, \bibinfo{author}{Thiha, P.},
  \bibinfo{author}{Zhong, Y.}, \bibinfo{author}{Brady, E.},
  \bibinfo{author}{Bigham, J.P.}, \bibinfo{year}{2013}.
\newblock \bibinfo{title}{Answering visual questions with conversational crowd
  assistants}, in: \bibinfo{booktitle}{Proceedings of the 15th International
  ACM SIGACCESS Conference on Computers and Accessibility},
  \bibinfo{publisher}{Association for Computing Machinery},
  \bibinfo{address}{New York, NY, USA}.
\bibitem[{Lau et~al.(2018)Lau, Gayen, Abacha and
  Demner-Fushman}]{lau2018dataset}
\bibinfo{author}{Lau, J.J.}, \bibinfo{author}{Gayen, S.},
  \bibinfo{author}{Abacha, A.B.}, \bibinfo{author}{Demner-Fushman, D.},
  \bibinfo{year}{2018}.
\newblock \bibinfo{title}{A dataset of clinically generated visual questions
  and answers about radiology images}.
\newblock \bibinfo{journal}{Scientific data} \bibinfo{volume}{5},
  \bibinfo{pages}{1--10}.
\bibitem[{{Li} et~al.(2019){Li}, {Zhang}, {Chen} and {Huang}}]{8510891}
\bibinfo{author}{{Li}, D.}, \bibinfo{author}{{Zhang}, Z.},
  \bibinfo{author}{{Chen}, X.}, \bibinfo{author}{{Huang}, K.},
  \bibinfo{year}{2019}.
\newblock \bibinfo{title}{A richly annotated pedestrian dataset for person
  retrieval in real surveillance scenarios}.
\newblock \bibinfo{journal}{IEEE Transactions on Image Processing}
  \bibinfo{volume}{28}, \bibinfo{pages}{1575--1590}.
\bibitem[{Li et~al.(2019)Li, Zhang, Yu, Huang and Tan}]{li2019isee}
\bibinfo{author}{Li, D.}, \bibinfo{author}{Zhang, Z.}, \bibinfo{author}{Yu,
  K.}, \bibinfo{author}{Huang, K.}, \bibinfo{author}{Tan, T.},
  \bibinfo{year}{2019}.
\newblock \bibinfo{title}{Isee: An intelligent scene exploration and evaluation
  platform for large-scale visual surveillance}.
\newblock \bibinfo{journal}{IEEE Transactions on Parallel and Distributed
  Systems} \bibinfo{volume}{30}, \bibinfo{pages}{2743--2758}.
\bibitem[{Liu et~al.(2019)Liu, Peng and Rosen}]{liu2019effective}
\bibinfo{author}{Liu, F.}, \bibinfo{author}{Peng, Y.}, \bibinfo{author}{Rosen,
  M.P.}, \bibinfo{year}{2019}.
\newblock \bibinfo{title}{An effective deep transfer learning and information
  fusion framework for medical visual question answering}, in:
  \bibinfo{booktitle}{International Conference of the Cross-Language Evaluation
  Forum for European Languages}, \bibinfo{organization}{Springer}. pp.
  \bibinfo{pages}{238--247}.
\bibitem[{{Liu} et~al.(2020){Liu}, {Xiang}, {Hospedales}, {Yang} and
  {Sun}}]{8528867}
\bibinfo{author}{{Liu}, F.}, \bibinfo{author}{{Xiang}, T.},
  \bibinfo{author}{{Hospedales}, T.M.}, \bibinfo{author}{{Yang}, W.},
  \bibinfo{author}{{Sun}, C.}, \bibinfo{year}{2020}.
\newblock \bibinfo{title}{Inverse visual question answering: A new benchmark
  and vqa diagnosis tool}.
\newblock \bibinfo{journal}{IEEE Transactions on Pattern Analysis and Machine
  Intelligence} \bibinfo{volume}{42}, \bibinfo{pages}{460--474}.
\bibitem[{Lubna et~al.(2019)Lubna, Kalady and Lijiya}]{lubna2019mobvqa}
\bibinfo{author}{Lubna, A.}, \bibinfo{author}{Kalady, S.},
  \bibinfo{author}{Lijiya, A.}, \bibinfo{year}{2019}.
\newblock \bibinfo{title}{Mobvqa: A modality based medical image visual
  question answering system}, in: \bibinfo{booktitle}{TENCON 2019-2019 IEEE
  Region 10 Conference (TENCON)}, \bibinfo{organization}{IEEE}. pp.
  \bibinfo{pages}{727--732}.
\bibitem[{Manmadhan and Kovoor(2020)}]{manmadhan2020visual}
\bibinfo{author}{Manmadhan, S.}, \bibinfo{author}{Kovoor, B.C.},
  \bibinfo{year}{2020}.
\newblock \bibinfo{title}{Visual question answering: a state-of-the-art
  review}.
\newblock \bibinfo{journal}{Artificial Intelligence Review} ,
  \bibinfo{pages}{1--41}.
\bibitem[{Mu{\~n}oz et~al.(2006)Mu{\~n}oz, Arellano, Perales and
  Fontanet}]{munoz2006perceptual}
\bibinfo{author}{Mu{\~n}oz, C.}, \bibinfo{author}{Arellano, D.},
  \bibinfo{author}{Perales, F.J.}, \bibinfo{author}{Fontanet, G.},
  \bibinfo{year}{2006}.
\newblock \bibinfo{title}{Perceptual and intelligent domotic system for
  disabled people}, in: \bibinfo{booktitle}{Proceedings of the 6th IASTED
  International Conference on Visualization, Imaging and Image Processing}, pp.
  \bibinfo{pages}{70--75}.
\bibitem[{Nguyen et~al.(2019)Nguyen, Do, Nguyen, Do, Tjiputra and
  Tran}]{nguyen2019overcoming}
\bibinfo{author}{Nguyen, B.D.}, \bibinfo{author}{Do, T.T.},
  \bibinfo{author}{Nguyen, B.X.}, \bibinfo{author}{Do, T.},
  \bibinfo{author}{Tjiputra, E.}, \bibinfo{author}{Tran, Q.D.},
  \bibinfo{year}{2019}.
\newblock \bibinfo{title}{Overcoming data limitation in medical visual question
  answering}, in: \bibinfo{booktitle}{International Conference on Medical Image
  Computing and Computer-Assisted Intervention},
  \bibinfo{organization}{Springer}. pp. \bibinfo{pages}{522--530}.
\bibitem[{Park et~al.(2019)Park, Lee, Kwon, Ha, Kim and Zhang}]{park2019ads}
\bibinfo{author}{Park, K.W.}, \bibinfo{author}{Lee, J.}, \bibinfo{author}{Kwon,
  S.}, \bibinfo{author}{Ha, J.W.}, \bibinfo{author}{Kim, K.M.},
  \bibinfo{author}{Zhang, B.T.}, \bibinfo{year}{2019}.
\newblock \bibinfo{title}{Which ads to show? advertisement image assessment
  with auxiliary information via multi-step modality fusion}.
\newblock \bibinfo{journal}{arXiv preprint arXiv:1910.02358} .
\bibitem[{Peng et~al.(2018)Peng, Liu and Rosen}]{peng2018umass}
\bibinfo{author}{Peng, Y.}, \bibinfo{author}{Liu, F.}, \bibinfo{author}{Rosen,
  M.P.}, \bibinfo{year}{2018}.
\newblock \bibinfo{title}{Umass at imageclef medical visual question answering
  (med-vqa) 2018 task.}, in: \bibinfo{booktitle}{CLEF (Working Notes)}.
\bibitem[{Ren and Zhou(2020)}]{ren2020cgmvqa}
\bibinfo{author}{Ren, F.}, \bibinfo{author}{Zhou, Y.}, \bibinfo{year}{2020}.
\newblock \bibinfo{title}{Cgmvqa: A new classification and generative model for
  medical visual question answering}.
\newblock \bibinfo{journal}{IEEE Access} \bibinfo{volume}{8},
  \bibinfo{pages}{50626--50636}.
\bibitem[{Ronchi and Perona(2015)}]{BMVC2015_52}
\bibinfo{author}{Ronchi, M.R.}, \bibinfo{author}{Perona, P.},
  \bibinfo{year}{2015}.
\newblock \bibinfo{title}{Describing common human visual actions in images},
  in: \bibinfo{editor}{Xianghua~Xie, M.W.J.}, \bibinfo{editor}{Tam, G.K.L.}
  (Eds.), \bibinfo{booktitle}{Proceedings of the British Machine Vision
  Conference (BMVC)}, \bibinfo{publisher}{BMVA Press}. pp.
  \bibinfo{pages}{52.1--52.12}.
\bibitem[{Sampat et~al.(2020)Sampat, Yang and Baral}]{sampat2020diverse}
\bibinfo{author}{Sampat, S.}, \bibinfo{author}{Yang, Y.},
  \bibinfo{author}{Baral, C.}, \bibinfo{year}{2020}.
\newblock \bibinfo{title}{Diverse visuo-lingustic question answering (dvlqa)
  challenge}.
\newblock \bibinfo{journal}{arXiv preprint arXiv:2005.00330} .
\bibitem[{Sheng et~al.(2016)Sheng, Gool and Moens}]{Sheng2016ADF}
\bibinfo{author}{Sheng, S.}, \bibinfo{author}{Gool, L.V.},
  \bibinfo{author}{Moens, M.F.}, \bibinfo{year}{2016}.
\newblock \bibinfo{title}{A dataset for multimodal question answering in the
  cultural heritage domain}, in: \bibinfo{booktitle}{LT4DH@COLING}.
\bibitem[{Shi et~al.(2019)Shi, Liu and Rosen}]{shi2019deep}
\bibinfo{author}{Shi, L.}, \bibinfo{author}{Liu, F.}, \bibinfo{author}{Rosen,
  M.P.}, \bibinfo{year}{2019}.
\newblock \bibinfo{title}{Deep multimodal learning for medical visual question
  answering}.
\newblock \bibinfo{journal}{Working Notes of CLEF} .
\bibitem[{Stefanini et~al.(2019)Stefanini, Cornia, Baraldi, Corsini and
  Cucchiara}]{10.1007/978-3-030-30645-8_66}
\bibinfo{author}{Stefanini, M.}, \bibinfo{author}{Cornia, M.},
  \bibinfo{author}{Baraldi, L.}, \bibinfo{author}{Corsini, M.},
  \bibinfo{author}{Cucchiara, R.}, \bibinfo{year}{2019}.
\newblock \bibinfo{title}{Artpedia: A new visual-semantic dataset with visual
  and contextual sentences in the artistic domain}, in: \bibinfo{editor}{Ricci,
  E.}, \bibinfo{editor}{Rota~Bul{\`o}, S.}, \bibinfo{editor}{Snoek, C.},
  \bibinfo{editor}{Lanz, O.}, \bibinfo{editor}{Messelodi, S.},
  \bibinfo{editor}{Sebe, N.} (Eds.), \bibinfo{booktitle}{ICIAP 2019},
  \bibinfo{publisher}{Springer International Publishing},
  \bibinfo{address}{Cham}. pp. \bibinfo{pages}{729--740}.
\bibitem[{Teney and Hengel(2019)}]{teney2019actively}
\bibinfo{author}{Teney, D.}, \bibinfo{author}{Hengel, A.v.d.},
  \bibinfo{year}{2019}.
\newblock \bibinfo{title}{Actively seeking and learning from live data}, in:
  \bibinfo{booktitle}{Proceedings of the IEEE Conference on Computer Vision and
  Pattern Recognition}, pp. \bibinfo{pages}{1940--1949}.
\bibitem[{Toor et~al.(2019a)Toor, Wechsler and Nappi}]{TOOR2019111}
\bibinfo{author}{Toor, A.S.}, \bibinfo{author}{Wechsler, H.},
  \bibinfo{author}{Nappi, M.}, \bibinfo{year}{2019}a.
\newblock \bibinfo{title}{Biometric surveillance using visual question
  answering}.
\newblock \bibinfo{journal}{Pattern Recognition Letters} \bibinfo{volume}{126},
  \bibinfo{pages}{111 -- 118}.
\newblock \bibinfo{note}{Robustness, Security and Regulation Aspects in Current
  Biometric Systems}.
\bibitem[{Toor et~al.(2019b)Toor, Wechsler and Nappi}]{toor2019question}
\bibinfo{author}{Toor, A.S.}, \bibinfo{author}{Wechsler, H.},
  \bibinfo{author}{Nappi, M.}, \bibinfo{year}{2019}b.
\newblock \bibinfo{title}{Question action relevance and editing for visual
  question answering}.
\newblock \bibinfo{journal}{Multimedia Tools and Applications}
  \bibinfo{volume}{78}, \bibinfo{pages}{2921--2935}.
\bibitem[{{Tu} et~al.(2014){Tu}, {Meng}, {Lee}, {Choe} and {Zhu}}]{6818956}
\bibinfo{author}{{Tu}, K.}, \bibinfo{author}{{Meng}, M.},
  \bibinfo{author}{{Lee}, M.W.}, \bibinfo{author}{{Choe}, T.E.},
  \bibinfo{author}{{Zhu}, S.}, \bibinfo{year}{2014}.
\newblock \bibinfo{title}{Joint video and text parsing for understanding events
  and answering queries}.
\newblock \bibinfo{journal}{IEEE MultiMedia} \bibinfo{volume}{21},
  \bibinfo{pages}{42--70}.
\bibitem[{Vu et~al.(2019)Vu, Sznitman, Nyholm and
  L{\"o}fstedt}]{vu2019ensemble}
\bibinfo{author}{Vu, M.}, \bibinfo{author}{Sznitman, R.},
  \bibinfo{author}{Nyholm, T.}, \bibinfo{author}{L{\"o}fstedt, T.},
  \bibinfo{year}{2019}.
\newblock \bibinfo{title}{Ensemble of streamlined bilinear visual question
  answering models for the imageclef 2019 challenge in the medical domain}, in:
  \bibinfo{booktitle}{CLEF 2019}.
\bibitem[{Vu et~al.(2020)Vu, L{\"o}fstedt, Nyholm and
  Sznitman}]{vu2020question}
\bibinfo{author}{Vu, M.H.}, \bibinfo{author}{L{\"o}fstedt, T.},
  \bibinfo{author}{Nyholm, T.}, \bibinfo{author}{Sznitman, R.},
  \bibinfo{year}{2020}.
\newblock \bibinfo{title}{A question-centric model for visual question
  answering in medical imaging}.
\newblock \bibinfo{journal}{IEEE Transactions on Medical Imaging} .
\bibitem[{Wang et~al.(2020)Wang, Huang and Wang}]{wang2020long}
\bibinfo{author}{Wang, W.}, \bibinfo{author}{Huang, Y.}, \bibinfo{author}{Wang,
  L.}, \bibinfo{year}{2020}.
\newblock \bibinfo{title}{Long video question answering: A matching-guided
  attention model}.
\newblock \bibinfo{journal}{Pattern Recognition} \bibinfo{volume}{102},
  \bibinfo{pages}{107248}.
\bibitem[{Weiss et~al.(2019)Weiss, Chamorro, Girgis, Luck, Kahou, Cohen,
  Nowrouzezahrai, Precup, Golemo and Pal}]{weiss2019navigation}
\bibinfo{author}{Weiss, M.}, \bibinfo{author}{Chamorro, S.},
  \bibinfo{author}{Girgis, R.}, \bibinfo{author}{Luck, M.},
  \bibinfo{author}{Kahou, S.E.}, \bibinfo{author}{Cohen, J.P.},
  \bibinfo{author}{Nowrouzezahrai, D.}, \bibinfo{author}{Precup, D.},
  \bibinfo{author}{Golemo, F.}, \bibinfo{author}{Pal, C.},
  \bibinfo{year}{2019}.
\newblock \bibinfo{title}{Navigation agents for the visually impaired: A
  sidewalk simulator and experiments}.
\newblock \bibinfo{journal}{arXiv preprint arXiv:1910.13249} .
\bibitem[{Wu et~al.(2017)Wu, Teney, Wang, Shen, Dick and van~den
  Hengel}]{wu2017visual}
\bibinfo{author}{Wu, Q.}, \bibinfo{author}{Teney, D.}, \bibinfo{author}{Wang,
  P.}, \bibinfo{author}{Shen, C.}, \bibinfo{author}{Dick, A.},
  \bibinfo{author}{van~den Hengel, A.}, \bibinfo{year}{2017}.
\newblock \bibinfo{title}{Visual question answering: A survey of methods and
  datasets}.
\newblock \bibinfo{journal}{Computer Vision and Image Understanding}
  \bibinfo{volume}{163}, \bibinfo{pages}{21--40}.
\bibitem[{Yan et~al.(2019)Yan, Li, Xie, Xiao and Gu}]{yan2019zhejiang}
\bibinfo{author}{Yan, X.}, \bibinfo{author}{Li, L.}, \bibinfo{author}{Xie, C.},
  \bibinfo{author}{Xiao, J.}, \bibinfo{author}{Gu, L.}, \bibinfo{year}{2019}.
\newblock \bibinfo{title}{Zhejiang university at imageclef 2019 visual question
  answering in the medical domain}.
\newblock \bibinfo{journal}{Working Notes of CLEF} .
\bibitem[{Zhang et~al.(2019)Zhang, Cao and Wu}]{zhang2019information}
\bibinfo{author}{Zhang, D.}, \bibinfo{author}{Cao, R.}, \bibinfo{author}{Wu,
  S.}, \bibinfo{year}{2019}.
\newblock \bibinfo{title}{Information fusion in visual question answering: A
  survey}.
\newblock \bibinfo{journal}{Information Fusion} \bibinfo{volume}{52},
  \bibinfo{pages}{268--280}.
\bibitem[{Zhou et~al.(2018)Zhou, Kang and Ren}]{zhou2018employing}
\bibinfo{author}{Zhou, Y.}, \bibinfo{author}{Kang, X.}, \bibinfo{author}{Ren,
  F.}, \bibinfo{year}{2018}.
\newblock \bibinfo{title}{Employing inception-resnet-v2 and bi-lstm for medical
  domain visual question answering.}, in: \bibinfo{booktitle}{CLEF (Working
  Notes)}.
\bibitem[{Zhou et~al.(2020)Zhou, Mishra, Verma, Bhamidipati and
  Wang}]{zhou2020recommending}
\bibinfo{author}{Zhou, Y.}, \bibinfo{author}{Mishra, S.},
  \bibinfo{author}{Verma, M.}, \bibinfo{author}{Bhamidipati, N.},
  \bibinfo{author}{Wang, W.}, \bibinfo{year}{2020}.
\newblock \bibinfo{title}{Recommending themes for ad creative design via
  visual-linguistic representations}, in: \bibinfo{booktitle}{Proceedings of
  The Web Conference 2020}, pp. \bibinfo{pages}{2521--2527}.

\end{thebibliography}

\end{document}